\documentclass[11pt,a4paper]{article}
\usepackage[hyperref]{acl2020}
\usepackage{times}
\usepackage{latexsym}

\usepackage{graphicx}
\usepackage{textcomp}
\usepackage{xcolor}
\usepackage{booktabs}
\usepackage{float}
\graphicspath{ {./} }
\usepackage{lipsum}% http://ctan.org/pkg/lipsum
\usepackage{multicol}% http://ctan.org/pkg/multicols

\usepackage{microtype}

\aclfinalcopy % Uncomment this line for the final submission
\title{Sentiment is all you need to win US Presidential elections}

\author{Sovesh Mohapatra\thanks{$^{*}$ Authors contributed equally} \\
  University of Massachusetts Amherst \\
  \texttt{soveshmohapa@umass.edu} \\\And
  Somesh Mohapatra$^{*}$ \\
  Massachusetts Institute of Technology \\
  \texttt{someshm@mit.edu} \\}

\date{}

\begin{document}

\maketitle
\begin{abstract}

Election speeches play an integral role in communicating the vision and mission of the candidates. From lofty promises to mud-slinging, the electoral candidate accounts for all. However, there remains an open question about what exactly wins over the voters. In this work, we used state-of-the-art natural language processing methods to study the speeches and sentiments of the Republican candidates and Democratic candidates fighting for the 2020 US Presidential election. Comparing the racial dichotomy of the United States, we analyze what led to the victory and defeat of the different candidates. We believe this work will inform the election campaigning strategy and provide a basis for communicating to diverse crowds.
\end{abstract}

\section{Introduction}

In a democracy, elections serve as the people's mandate. They speak for what the people think and who they want to represent their voices. However, the mandate is not without bias and is often swayed by the communication at election rallies, on social media, and at dinner table conversations \citep{anderson2001winners}. Understanding what reinforces people's opinions, or changes them, is a complex question \citep{johnston1992letting}.

Election campaign of a candidate attempts to decode what the people want and focuses the messaging around that \citep{dupont2019kind}. The audience strategy is usually based on heuristics like the assumption that voters vote based on their economic interests leads them to change the economy for the better, for instance, by introducing support funds or creating new jobs, especially shortly before an election \citep{powell1993cross, whitten1999cross}. These are also being dominated by people who happen to know the intricacies of the ground \citep{mcclurg2004indirect}. With the rise in usage of \textit{post hoc} analysis of strategies, elections have started to become less heuristic-driven and more data-driven \citep{anstead2017data}. One of the most prominent and mobilizing parts of the elections are the speeches, and the use of language, by the running candidates \citep{steffens2013power,ikeanyibe2018political}. The sentiments and statements in speeches are often described as an art.

In this paper, we attempted to decode the art of speeches by focusing on the sentiment classification of speeches across various states and demographics, and how their effect on the election results. A survey was also conducted to analyze people's responses to snippets of the speeches.

\section{Related Work}
In the social sciences field, there have been multiple approaches to analyze the US Presidential election speeches. Populism framing in the speeches has been explored to analyze speeches in a novel database comprising of speeches from 1896 to 2016 in \citet{fahey2021building}, and a more recent focus with active metaphors in \citet{keating2021populist}. Political rhetoric \cite{bull2015whipping, conway2012does}, deception strategies \cite{al2017pragmatics}, and metrical analysis \cite{ban2009metrical}, amongst other linguistic approaches have been used to study Presidential elections for different years. However, these approaches do not leverage the significant advances in artificial intelligence to push forth their analyses.

The rise of sentimental analysis and widespread availability of Twitter data have contributed to more computational analysis in the recent years. Several studies analyzed the Twitter responses, tweets by Donald Trump, and their aftermath using NLP tools \cite{liu2017reviewing, yaqub2017analysis, caetano2018using, siegel2021trumping}. Political sentiment analysis and the use of sentiment analysis to predict election results has been attempted \cite{nausheen2018sentiment, elghazaly2016political, liu2018appeal}. \citet{finity2021text} provide a text analysis of the 2020 US election speeches. We believe that the extension of these approaches to more recent emnotion-based approaches, along with human surveys could provide a robust method of understanding the effect that the speeches have.

In 2021, GoEmotions, a database of fine-grained emotions, labeled for 27 emotion categories was released by a team of researchers from Google, Amazon and Stanford Linguistics Department \cite{demszky2020goemotions}. Sequence to emotion models were developed in \citet{huang2021seq2emo}, visualization of the emotions was done in \citet{dumontvisualing}, and these models were applied to text sentiment analysis and essay analysis  \cite{thainguan2021text, maheshwari2022ensemble}. More advanced models, such as Emoroberta, have been recently developed \cite{kamath2022emoroberta}, and limitations of the text-based emotion detection have also been discussed \cite{alvarez2021uncovering}.

\section{Methodology}
\subsection{Data Collection and Processing}

We have collected the transcripts from 61 election rally speeches that Republican candidate Donald Trump had given in various states between 2018 and 2020. Considering the Democrats, we have collected around 85 rally speeches by Barack Obama, Kamala Harris, and Joe Biden. The transcripts of the speeches were web-scrapped from various online news and transcripts such as  \url{USNews.com} (accessed on May 15, 2022), \url{CNN.com} (accessed on June 10, 2022), and \url{Rev.com} (accessed on June 15, 2022).

Each speech was then classified based on the state they were delivered. After which, we labelled the states into two different race categories: Black or White states. The state's label was based on the statistics that in the US, 14.9\% identified as Black or African American from \url{blackdemographics.com}  (accessed on July 4, 2022).  So, a state with a population of more than 14.9\% of people identifying themselves as Black or African Americans is considered a Black state. Later, it was again categorized into four new categories: loss in White (White state where the party has lost the elections), win in White (White state where the party has won the elections), loss in Black (Black state where the party has lost the elections) and win in Black (Black state where the party has won the elections).

Further, we clustered all the classified White state's speeches into one and all the Black state's speeches into the other. Then, we tokenized each sentence and passed it through our fine-tuned BERT model to classify the different sentences into the twenty seven selected emotions.

\subsection{Human Survey Collection}

Along with the machine categorisation of the sentences in the speeches, we took fifteen sentences representing a mix of nine types of emotions. These fifteen sentences were snippets from the various speeches delivered by candidates of both parties. Out of fifteen snippets, nine snippets didn't have information about the speaker, and the remaining six snippets had information about who is the speaker of the snippet. Out of the six, the last two snippets presented with interchanging the speakers' names.
 (see Table~\ref{Example}). 

The survey was a digital form which had a geographical location question to understand the demographics of the people taking the survey and two questions per snippet: whether the individual would vote for the candidate by just listening to this snippet, and from which party the speaker of the snippet was. We collected 68 responses, with the age of people ranging from 18 to 60. All participants in the survey were randomly selected from a pool of professors, students and staffs from various departments of the university which helped in getting survey takers from various states of the United States. This helped in giving us the idea required for the geographic location specific ideologies that require the local government to govern in a better way.

\subsection{Model Training}
We used a BERT model, given its ability to provide context-dependent token-level representations from whole sentences \citep{devlin2018bert, suhr2018learning}, unlike word-by-word and context-independent GloVebased or Word2Vec embeddings \citep{miaschi2020contextual, dev2020oscar}.

\begin{table}
\caption{\label{Example} Examples of snippets used in the survey.}
\begin{tabular}{p{3.8cm}p{1.6cm}p{1.8cm}}
\hline \centering \textbf{Snippets from Speeches} & \textbf{Party} & \textbf{Emotion} \\
\hline
As I personally told the Taliban leader if anyone ever double crossed the USA it would be the last thing they ever did & Republican & Negative, Anger, Optimism \\
\\I’ll never forget what President Kennedy said about going to the moon & Democratic & Positive, Optimism \\
\\ Would be good to talk to him rather than nuclear war wouldn’t it be nice? Anyway through a series of events I did talk to him and it was nasty at the beginning remember & Republican & Positive, Optimism, Gratitude \\
\\Think about what it takes to be a Black person who loves America today & Democratic & Positive, Admiration, Love \\
\hline
\end{tabular}
\end{table}

\subsection{Fine-tuning of the Model}

The model's downstream performance is essential for activities for specific usages, such as sentimental analysis of human conversations, thereby requiring fine-tuning of generic models for specific actions \cite{devlin2018bert}. We fine-tuned the BERT model using the GoEmotions dataset, a corpus of sentences classified into 27 different emotions \citep{demszky2020goemotions}. Using a transfer learning strategy, the model parameters were updated by training over the GoEmotions labeled corpus for 25 epochs.

\section{Results and Discussion}
\label{sec:results}

\subsection{Republicans and Democrats use similar sentiments, on average}
\begin{figure*}[t]
\centering
\includegraphics[width=1\textwidth]{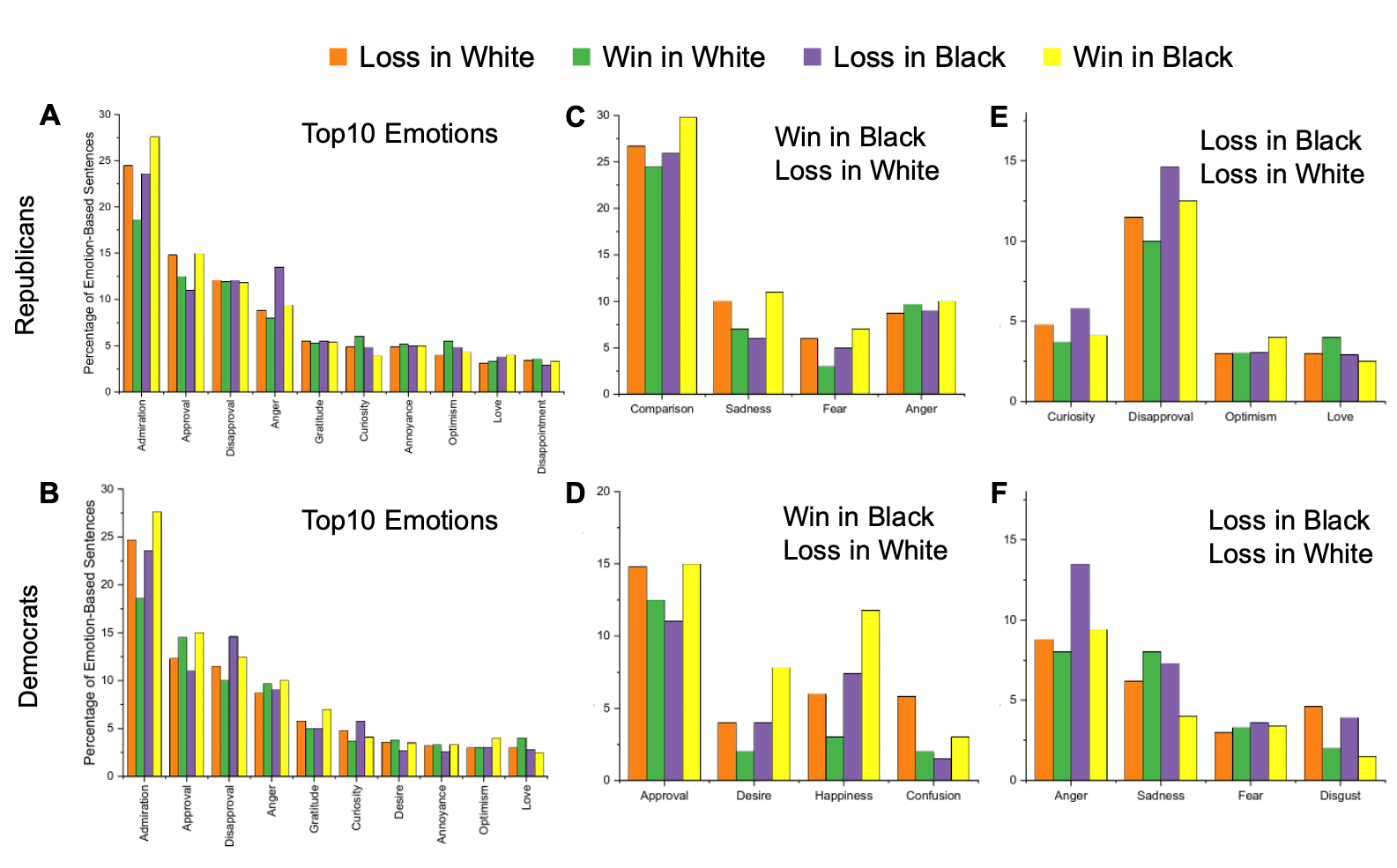}
\caption{Top 10 Emotions in the Speech by A. Republicans and B. Democrats. Emotions that led to win in Black and loss in White states for C. Republicans and D. Democrats; and loss in both states for E. Republicans and F. Democrats.}
\label{fig1}
\end{figure*}

We noted that the top 10 emotions (sentence-wise) used in the speeches in both White and Black states delivered by both parties follow a similar template, which could sometimes create a bias for the public in choosing their candidate (Figures \ref{fig1}A, \ref{fig1}B). This result also demonstrates how both parties attempt to use similar sentiments to attract voters.

\subsection{Comparison in Speech wins in Black states, loses in White states for Republicans}
\label{sec:gratitude}

The sentences with comparison and sadness emotions played a significant role in the speeches delivered in the Black states that were won by the Republicans (Figure \ref{fig1}C). However, they had to face a loss in the White states with a higher frequency of the same category. This difference shows the different aspirations of the White and Black population. Comparison includes disapproval, approval and confusion; while sadness clusters remorse, grief and disappointment emotions. 

\subsection{Approval and Desire in Speech win in Black states, lose in White states for Democrats}
\label{sec:desire}

The sentences based on emotions such as approval and happiness played a significant role in all the speeches delivered in the Black states that were won by the Democrats, in line with the liberal ideology (Figure \ref{fig1}D). In contrast, they had to face a loss in the White states with a higher frequency of the same category, owing to the strong Republican pull. Happiness clusters amusement, excitement and joy, and desire includes gratitude emotions. 

\subsection{Curiosity and Disapproval leads to loses in both Black and White states for Republicans}
\label{sec:curiosity}

The sentences categorized as curiosity and disapproval in speeches led the Republicans to lose in both Black and White states. Interestingly,  disapproval sentences are one of the top three kinds of sentences that Republicans used in their speeches (Figure \ref{fig1}E). It is also one of the top negative emotions used by the Republicans. This difference in the results may be attributed to the influence of the Democrats during the election campaigns, and the results coming in from the swing states.

\subsection{Anger and Disgust lead to loses in both Black and White states for Democrats}

\label{sec:anger}

The sentences categorized as anger and disgust in speeches led the Democrats to lose in both the Black and White states. As with the observation in the case of Democrats, anger sentences are one of the top four kinds of sentences used by Democrats. It is also one of the top negative emotions used by the Democrats. This anomaly shows us the expectation of the people from the Democrats not to show negative emotions like anger and disgust towards any matter and instead come up with a solution to dissolve the situation. (Figure \ref{fig1}F).

\subsection{Positivity wins in White states, loses in Black states for Republicans, and vice-versa for Democrats }
\label{sec:Positivity}

\begin{figure}
\centering
\includegraphics[width=0.4\textwidth]{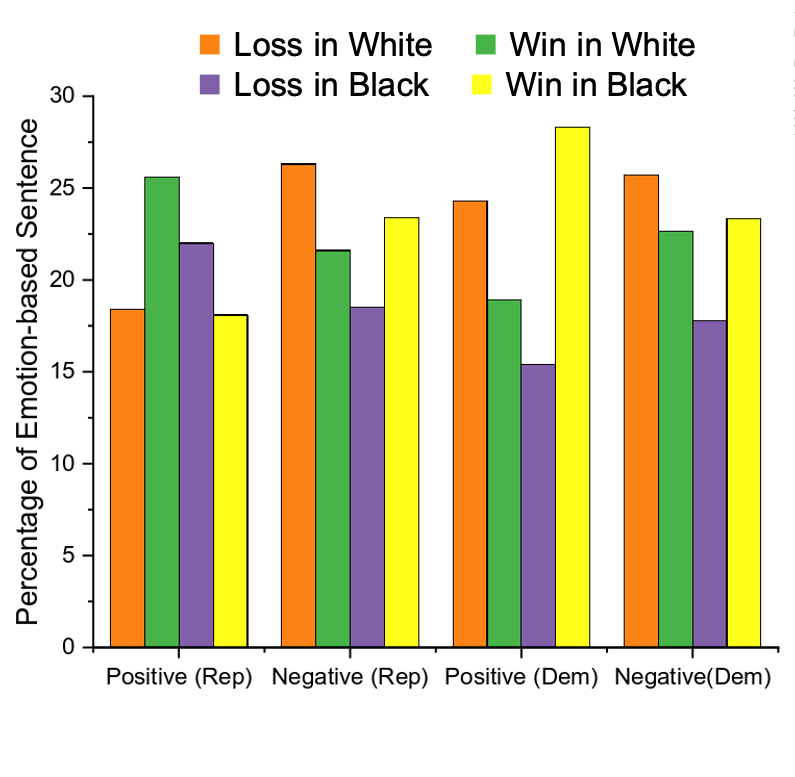}
\caption{Net impact of sentiments, positive and negative, for Republicans and Democrats.}
\label{fig2}
\end{figure}

We observed that when the overall notion of the speech was positive, it favored the Republicans and not the Democrats to win elections in the White states. In contrast, this is reversed when the overall notion of the speech becomes negative. The Republicans won the elections in the Black states, but the Democrats had to face loss (Figure \ref{fig2}). The reversal of notions in the speeches and wins in Black versus White states for Republicans and Democrats highlights their approach to the different demographies, and how it played out in the results.

The positive notion of a speech was calculated by clustering the frequencies of the positive emotions: gratitude, optimism, love, excitement, caring, joy, and amusement. Similarly, the negative notion of the speech was calculated by clustering the frequencies of negative emotions: annoyance, disappointment, anger, fear, sadness, disgust, and embarrassment.

\subsection{Survey Findings}

Along with observing various sentiments swaying the results of an election, we found that the sentiments are not the only factor behind deciding the influence of the speech. From the survey, we found the emotions like desire and happiness categorized sentences when given to the people by indicating to them that the snippet is from the Democrats, then the individual's choice to vote increased. In contrast, when a similar emotion-based snippet was given by blinding the information about the candidate, the individual opinion to vote varied. Similar results were observed when the emotion of curiosity-based snippets were provided that led the individual's choice to vote for Republicans decreased. However, when we blinded the information about the candidate, the similar emotion-based snippets got different opinions. This result explains why it is crucial to understand an individual's expectations from the candidate they hear to. 

From our machine categorization of snippets, we also saw that the emotion curiosity-based sentences were not also in favor of the Republicans when delivered in either Black or White states.

\section{Limitations, and Future Work}

The small lexicon considering just speeches targeting the 2020 Presidential elections results in detecting a relatively narrow understanding of how the sentiments used in the speeches affect the individuals. Furthermore, the idea behind voting for an individual comes with many prejudices against the candidate and the party representing. This was observed in the survey when we put forward the same snippet and attributed it to a different speaker. In our study, we nevertheless saw an interesting set of emotions that have driven the 2020 Presidential elections for either party, such as Section~\ref{sec:Positivity}, where we noted how positivity and negativity notions had impacts when used by either party, and Section~\ref{sec:curiosity} and~\ref{sec:anger}, where we noted that when the parties used particular emotion-based sentences, they had to face a loss.

Future work, including understanding the impact of emotions on voters from different backgrounds, such as immigrants, white- and blue-collar workers, and other demographics, can shed light on the relationship between how the particular emotion-based sentences can sway the elections. Considering multiple years of Presidential elections rally speeches and understanding the opinions biased by the individual's background would be vital in understanding the changing landscape of people's aspirations and how they are catered to by the candidates.  

\section{Conclusion}

We collected a large-scale political rally speech of the 2020 Presidential elections to understand how speeches and sentiments have influenced the opinion of people voting for a particular candidate. Our analysis confirmed that different kinds of emotion-based sentences sway people's views about voting for a candidate. In contrast, we also observed that people wanted to listen to a particular party about a specific topic using a set of emotion-based sentences. Our analysis demonstrated that if the emotion could be identified \textit{a priori} and delivered by a specific candidate, the election strategy could be targeted and aligned to the voters' bias. 

\bibliography{acl2020}
\bibliographystyle{acl_natbib}

\end{document}